# ChatGPT produces more "lazy" thinkers: Evidence of cognitive engagement decline

Georgios P. Georgiou[1] [2]

[1]Department of Languages and Literature, University of Nicosia

[2]Phonetic Lab, University of Nicosia

georgiou.georg@unic.ac.cy

**Abstract**

Despite the increasing use of large language models (LLMs) in education, concerns have emerged about their potential to reduce deep thinking and active learning. This study investigates the impact of generative artificial intelligence (AI) tools, specifically ChatGPT, on students' cognitive engagement during academic writing tasks. The study employed an experimental design with participants randomly assigned to either an AI-assisted (ChatGPT) or a non-assisted (control) condition. Participants completed a structured argumentative writing task followed by a cognitive engagement scale (CES), the CES-AI, developed to assess mental effort, attention, deep processing, and strategic thinking. The results revealed significantly lower cognitive engagement scores in the ChatGPT group compared to the control group. These findings suggest that AI assistance may lead to cognitive offloading. The study contributes to the growing body of literature on the psychological implications of AI in education and raises important questions about the integration of such tools into academic practice. It calls for pedagogical strategies that promote active, reflective engagement with AI-generated content to avoid compromising students' self-regulated learning and deep cognitive involvement.

**Introduction**

Recent breakthroughs in artificial intelligence (AI) and natural language processing (NLP) have led to the development of highly advanced language technologies known as Large Language Models (LLMs) (Georgiou, 2025). As a form of generative AI, LLMs are capable of producing original content by identifying and applying linguistic patterns learned from vast datasets. These models enable machines to understand and generate human-like language (Zhu et al., 2024). One prominent example is ChatGPT (short for Generative Pretrained Transformers), developed by OpenAI (OpenAI, 2023), which employs NLP techniques to generate contextually appropriate text in response to user input. As these tools become increasingly integrated into educational contexts, particularly in writing and language-based tasks (Imran & Almusharraf, 2023), questions are emerging about their cognitive implications. Specifically, concerns have been raised about whether the use of AI tools like ChatGPT might alter the way students mentally engage with learning activities. While the functional capabilities of LLMs are well documented, their effects on learners' cognitive engagement remain underexplored. This research gap is addressed by the present study, which employs an experimental approach to provide empirical insights into how the use of AI tools may influence students' cognitive engagement in language tasks.

The concept of student "engagement" has primarily been defined within school-based contexts. Fredricks et al. (2004) identified three distinct facets of engagement: behavioral, emotional, and cognitive. Regarding cognitive engagement, different definitions were provided in the literature. These definitions collectively describe the term as the mental investment and strategic thinking that students apply during learning. It involves motivation, self-regulation, goal-setting, and effortful processing of tasks. Some definitions emphasize broader school-level engagement and the relevance of learning to future goals (e.g., Appleton et al., 2006; Furlong & Christenson, 2008), while others focus on task-specific effort and autonomy (e.g., Helme & Clarke, 2001; Rotgans et al., 2011). Across sources, cognitive engagement consistently includes thoughtful involvement, psychological investment, and the use of learning strategies to understand and master material.

In the last few years, research has shifted toward investigating cognitive engagement, with a specific focus on "attrition" of cognitive skills in relation to the use of AI. For example, Kosmyna et al. (2025) investigated the cognitive costs of using LLMs in educational essay writing. Participants were assigned to one of three groups: LLM, Search Engine, or Brain-only (no tools), across three sessions. In a fourth session, some participants switched conditions to assess cognitive flexibility. EEG was used to measure cognitive engagement, complemented by NLP analysis, human and AI scoring, and post-task interviews. The results showed that reliance on external tools, especially LLMs, reduced neural connectivity, with the Brain-only group exhibiting the strongest cognitive engagement. Participants using LLMs showed weaker memory recall, less ownership of their work, and poorer neural and linguistic performance compared to other groups. In addition, session four revealed lingering effects of tool reliance, with LLM-to-Brain participants underperforming even after tool removal. The study highlights that while LLMs offer convenience, their extended use may undermine deep learning, memory, and engagement, raising concerns about their long-term impact on education.

This systematic review by Lo et al. (2024) analyzed empirical studies published within a year of ChatGPT's release to evaluate its influence on student engagement across behavioural, emotional, and cognitive dimensions. Drawing on prior SWOT analyses, the review found strong but narrowly focused evidence of behavioural engagement, particularly in students' active use of ChatGPT and concerns about academic dishonesty. Emotional engagement yielded mixed results, with students reporting both positive (e.g., satisfaction, enjoyment) and negative (e.g., anxiety, disappointment) responses. Evidence for cognitive engagement was broad but weak, including reports of improved understanding and self-perception, alongside risks like diminished critical thinking and overreliance on AI. The review highlights gaps in the literature, calling for future research to explore overlooked indicators, including critical thinking (i.e., cognitive engagement), among others.

This study investigates the impact of generative AI tools on students' cognitive engagement during academic writing tasks. While previous research has largely focused on the accuracy, usefulness, or ethical implications of AI in education (Chen et al., 2020; Mhlanga, 2023), few studies have examined how such tools affect students' deeper mental involvement in learning tasks. Uniquely positioned at the intersection of AI and educational and cognitive psychology, the study employed a controlled experimental design involving two groups of students: one that completed an essay task using ChatGPT, and another that completed the same task without AI assistance. Following the writing task, all participants completed a cognitive engagement scale (CES) to assess their self-perceived levels of mental effort, strategic thinking, and investment in the task. By directly comparing cognitive engagement between the AI-assisted and non-assisted groups, this study offers novel insights into how AI use may alter students' depth of thinking and learning behavior. The findings contribute to a growing body of research on the educational and psychological implications of AI, shedding light on whether generative AI tools support or undermine students' cognitive engagement in high-level academic tasks.

## Methodology

### Participants

The study recruited 40 students, aged between 25 and 47 years ($M$age $=$ 35.12, $SD =$ 6.18). Gender representation was balanced to control for potential gender-related differences in cognitive engagement and familiarity with AI tools. Participants were either current or former students of programs specializing in linguistics, applied linguistics, and related language fields. All participants were native Greek speakers with at least a bachelor's degree, ensuring sufficient academic background to comprehend the task and questionnaire. Eligibility criteria included basic familiarity with AI chatbots such as ChatGPT (self-reported use of 16.4 hours per week; with an $SD$ of 5.93), no history of neurological or psychiatric disorders, and normal or corrected-to-normal vision.

Participants provided informed consent in accordance with ethical guidelines approved by the institutional review board. Upon enrollment, they were randomly assigned to either the ChatGPT group or the control group, with 20 participants in each. The two groups did not differ in terms of AI use [$t = -0.74$, df = 36.72, $p$-value = 0.46].

**Instrument**

The primary task required participants to write a structured argument either supporting or opposing the statement, "Educational institutions should integrate AI tools into standard academic practice". This prompt was selected because argumentative writing is cognitively demanding, requiring critical thinking, planning, and synthesis of ideas. Participants were instructed to write at least 300 words and were allotted a maximum of 30 minutes to complete the task. This writing task was adapted from standard argumentative prompts commonly used in cognitive engagement and educational research.

Following the writing task, participants completed a CES, a four-item self-report measure developed specifically for this study to assess mental effort and involvement during the task, especially in the context of ChatGPT use. Although selfreports might introduce bias, they are still widely used and have a position in psychological research (Greene, 2015). Since there are no standard CESs (Li, 2021), we constructed a scale with four items, which we call CES-AI. Each item was rated on a five-point Likert scale ranging from 1 (Strongly Disagree) to 5 (Strongly Agree). Higher average scores indicated greater cognitive engagement. The items of CES-AI measured facets of deep processing, effort, attention, and strategic engagement. The scale's internal consistency was evaluated using Cronbach's alpha. The resulting alpha coefficient was 0.88, which is considered adequate according to established thresholds (Taber, 2018). This value suggests that the four items consistently measure the same underlying construct – cognitive engagement – and that the scale demonstrates strong internal reliability. The items are presented in Table 1.

Table 1: Items of the CES-AI.

| Number | Item |
| --- | --- |
| 1 | I tried to understand the task deeply, rather than just skim through it. |
| 2 | I put effort into thinking through the problem myself. |
| 3 | I stayed mentally focused throughout the task. |
| 4 | I explored different ways to solve the problem or approach the task. |

As mentioned earlier, currently, no validated self-report scales exist measuring cognitive engagement in the context of AI tool use. This gap highlights the need to adapt or develop tailored instruments that can accurately capture the unique cognitive dynamics involved in AI-assisted learning tasks. The reasoning behind the development of the current scale items is presented as follows. Item 1 reflects deep processing and elaboration, a core feature of cognitive engagement (Fredricks et al., 2004), as the latter is about meaningful investment in understanding rather than surface-level processing. Item 2 reflects effortful thinking, which is fundamental to cognitive engagement (Greene, 2015; Schaufeli et al., 2002). This item captures active mental effort and self-regulated processing (Zimmerman, 2002). Item 3 relates to sustained attention, which is crucial to cognitive engagement, as distraction or mind-wandering reduces engagement (Smallwood & Schooler, 2006). This item taps the participants' ability to maintain concentration, since focus comprises a component of engagement and cognitive control. Along the same lines, Skinner et al. (2009) identified attention regulation as part of student engagement. Finally, item 4 signals strategic thinking and metacognition through the exploration of multiple strategies, which are essential for cognitive engagement (Pintrich, 2004); it shows flexibility and active involvement in learning.

**Procedure**

Participants were randomly assigned to one of two conditions: the ChatGPT condition or the control condition. In the ChatGPT condition, participants were allowed to use ChatGPT 3.5 during the reasoning task. They received instructions indicating that they could consult the AI tool for ideas, phrasing, or argument development, but were encouraged to engage actively and not rely solely on the AI's suggestions. In the control condition, participants were instructed to complete the reasoning task independently without using any external help or AI tools.

Participants completed the writing task on an online experimental platform that provided a text input box with a visible word count. They were given a maximum of 30 minutes to complete their response, but could submit earlier if they finished. All students were monitored via TeamViewer, with their cameras on and screens shared, to ensure that neither group received outside assistance. This also ensured that the control group did not use AI during the writing task, while the ChatGPT group did. Immediately after submitting their written response, participants completed the CES-AI, which was presented in random order to minimize response bias. Finally, participants answered demographic questions including age, gender, highest education level completed, and their frequency of prior use of ChatGPT or other AI chatbots. All data were collected anonymously and securely stored. The dataset included the written texts, CES-AI responses, and demographic information.

**Results**

The descriptive results demonstrated that the experimental group (i.e., those who used ChatGPT) scored lower ($M = 2.95$, $SD = 1.18$) than the control group (i.e., those who did not use ChatGPT) ($M = 4.19$, $SD = 0.45$) in the CES-AI. Figure 1 displays the performance of the experimental and control groups on the scale.

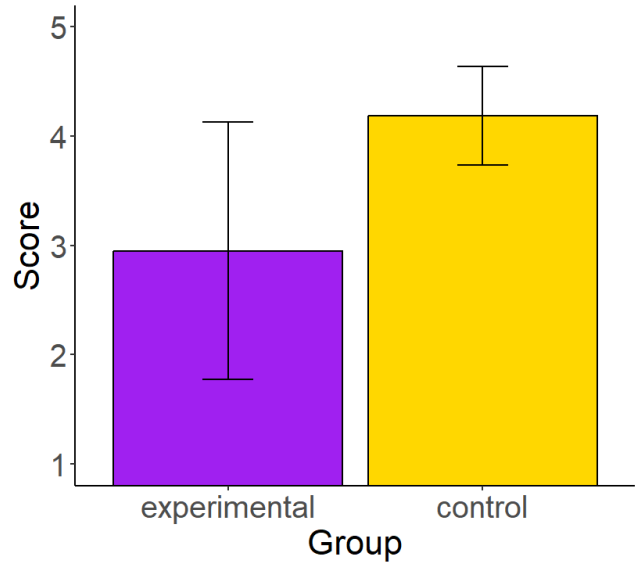

Figure 1: Scores (Likert-point scale from 1–5) of the experimental and control groups in the CES-AI.

To examine whether this difference was statistically significant, we used a one-way ANOVA test in R software (R Core Team, 2025). Score, which included the average score of each participant across the four items of the CES-AI, was modelled as the dependent variable, and Group (experimental/control) was modeled as the independent variable. The results showed that there was a significant effect of Group on Score [$F(1,38) = 19.2$, $p < 0.001$]. This finding indicated that the controlled group exhibited significantly higher cognitive engagement scores compared to the experimental group.

**Discussion**

The present study investigated students' cognitive engagement during academic writing tasks under two conditions: one involving the use of ChatGPT and one without. Cognitive engagement was assessed immediately following the task using a newly developed instrument, CES-AI, which was designed for this context.

The results demonstrated that students who did not use ChatGPT during the writing task reported significantly higher levels of cognitive engagement compared to those who completed the task with AI assistance. Specifically, participants who did not use ChatGPT were more likely to report engaging in deep processing, reflecting a stronger commitment to understanding the task beyond surface-level comprehension. They also indicated exerting greater mental effort, suggesting higher levels of self-regulated thinking. Furthermore, these students reported stronger sustained attention, meaning they were better able to maintain focus throughout the task even in the presence of digital tools. Finally, the absence of ChatGPT use appeared to support metacognitive engagement, with participants demonstrating greater strategic flexibility by exploring multiple approaches to the problem. These findings suggest that, rather than enhancing mental effort, AI tools like ChatGPT may reduce learners' cognitive engagement when thoughtfully integrated into academic tasks. These findings align with previous research in the literature. For example, Kosmyna et al. (2025) reported that the use of LLMs diminished deep learning and user engagement. Similarly, the systematic review by Lo et al. (2024) highlighted reduced cognitive engagement following the use of ChatGPT. The unique contribution of the present study lies in its extension of this line of work through the application of a CES to quantify cognitive engagement.

Contrary to some expectations that AI assistance might enhance cognitive effort by providing scaffolding (e.g., Liao et al., 2024), the findings suggest a potential cognitive offloading effect, where reliance on AI corresponds with reduced cognitive engagement. These results support theoretical models that view cognitive engagement as multifaceted (Lin et al., 2023), involving active investment and self-regulation, which may be compromised when learners delegate critical thinking to AI tools. The findings indicate the need to refine engagement theories to account for emerging digital learning contexts where human-AI interaction dynamically shapes motivation and effort. From a practical teaching perspective, the study signals caution in the uncritical adoption of AI tools like ChatGPT in educational settings. Educators should be aware that while these tools can support content generation and idea exploration, they may inadvertently decrease students' cognitive engagement, potentially undermining deep learning and self-regulation skills crucial for academic success (Panadero et al., 2021). Pedagogical strategies should focus on designing AI-assisted tasks that promote active learner involvement, such as prompting students to critically evaluate AI-generated content, encouraging metacognitive reflection alongside AI use. Instructors might also consider blended approaches where AI tools complement rather than replace cognitive effort, fostering a balance between technology assistance and learner autonomy.

## Conclusions

Together, the results suggest that AI tools may encourage students to become "lazy thinkers", potentially reducing their cognitive engagement when relying heavily on tools like ChatGPT. While this finding is important, several limitations must be acknowledged. First, the study relied exclusively on selfreport measures to assess cognitive engagement. Although selfreports provide valuable insight into participants' subjective experiences, they may not fully capture the actual depth or quality of cognitive engagement during the task. Factors such as social desirability bias, inaccurate self-perception, or a lack of awareness about one's own cognitive processes can limit the validity of selfreported data. Furthermore, cognitive engagement is a complex construct that may benefit from being measured through multiple complementary methods. Future research could improve upon this by triangulating selfreport data with objective measures such as neurophysiological recordings, behavioral indicators, or qualitative data from think-aloud protocols and interviews. Combining these approaches would offer a more comprehensive understanding of how AI tools affect students' cognitive involvement. Moreover, the present study's sample size and context may limit the generalizability of the findings. Larger and more diverse participant groups, as well as varied academic tasks and AI tools, should be explored in subsequent research to validate and extend these results.

## Acknowledgements

This study was supported by the Phonetic Lab of the University of Nicosia.